\documentclass[preprints,article,accept,moreauthors,pdftex,10pt,a4paper]{mdpi}

\firstpage{1} 
\pubvolume{xx}
\issuenum{1}
\articlenumber{1}
\pubyear{2018}
\copyrightyear{2018}
\externaleditor{Academic Editor: name}
\history{Received: date; Accepted: date; Published: date}
\Title{A Non-structural Representation Scheme for Articulated Shapes}

\Author{Asli Genctav $^{1}$* and Sibel Tari $^{1}$}

\AuthorNames{Asli Genctav and Sibel Tari}

\address{%
$^{1}$ \quad Department of Computer Engineering, Middle East Technical University, 06800, Ankara, Turkey}

\corres{Correspondence: asli@ceng.metu.edu.tr}

\abstract{For representing articulated shapes, as an alternative to the structured models based on graphs representing part hierarchy, we propose a pixel-based distinctness measure. Its spatial distribution yields a partitioning of the shape into a set of regions each of which is represented via size normalized probability distribution of the distinctness. Without imposing any structural relation among parts, pairwise shape similarity is formulated as the cost of an optimal assignment between respective regions. The matching is performed via Hungarian algorithm permitting some unmatched regions. The proposed similarity measure is employed in the context of clustering a set of shapes. The clustering results obtained on three articulated shape datasets show that our method performs comparable to state of the art methods utilizing component graphs or trees even though we are not explicitly modeling component relations.}


\begin{document}
%%%%%%%%%%%%%%%%%%%%%%%%%%%%%%%%%%%%%%%%%%
%% Only for the journal Gels: Please place the Experimental Section after the Conclusions

%%%%%%%%%%%%%%%%%%%%%%%%%%%%%%%%%%%%%%%%%%
%\setcounter{section}{-1} %% Remove this when starting to work on the template.
%\section{How to Use this Template}
%The template details the sections that can be used in a manuscript. Note that the order and names of article sections may differ from the requirements of the journal (e.g. the positioning of the Materials and Methods section). Please check the instructions for authors page of the journal to verify the correct order and names. For any questions, please contact the editorial office of the journal or support@mdpi.com. For LaTeX related questions please contact Janine Daum at latex-support@mdpi.com.
%The order of the section titles is: Introduction, Materials and Methods, Results, Discussion, Conclusions for these journals: aerospace,algorithms,antibodies,antioxidants,atmosphere,axioms,biomedicines,carbon,crystals,designs,diagnostics,environments,fermentation,fluids,forests,fractalfract,informatics,information,inventions,jfmk,jrfm,lubricants,neonatalscreening,neuroglia,particles,pharmaceutics,polymers,processes,technologies,viruses,vision

\section{Introduction}

Shape is a distinctive object attribute which is frequently utilized in image processing and computer vision applications.
Measuring similarity of objects via their shapes is a difficult task due to high within-class and low between-class variations.
Within-class variations may be due to transformations such as rotation, scaling and deformation of articulations.
Articulated shapes can be successfully represented by structural representations which are organized in the form of graphs of shape components.
However, it is challenging to build and compare structural representations.
%
%For example, in order to obtain a clean and consistent representation, skeleton extraction is frequently assisted by pruning which involves several heuristics.
%
Moreover, measuring similarity of shapes through their structural representations requires finding a correspondence between a pair of graphs, which is an intricate process entailing advanced algorithms.

In this work, we propose a representation scheme for articulated shapes which involves neither building a graph of shape components nor matching a pair of graphs.
The proposed representation is used to measure pairwise shape similarity according to which we cluster a set of shapes.
The clustering results obtained on three articulated shape datasets show that our method performs comparable to state of the art methods utilizing component graphs or trees even though we are not explicitly modeling component relations.

\section{The Method}

Our representation scheme relies on first constructing multiple high-dimensional feature spaces in which shape points (pixels in 2D discrete setting) are represented and then, determining distinctness of the shape points in each space separately via Robust Principal Component Analysis (RPCA).

The distinctness values deduced from each feature space are utilized for two main purposes.
First, their spatial distribution on the 2D shape domain is used to partition the shape into a set of regions.
Second, each region is described by the normalized probability distribution of the corresponding distinctness values.
The dissimilarity between a pair of shapes via each feature space is defined as the cost of the optimal assignment between their regions.
Notice that we do not build any graphs to model the shape structure and the optimal assignment problem does not involve matching a pair of graphs.
The final shape dissimilarity is computed by combining the dissimilarities deduced from multiple feature spaces.
Below, we present the details of our representation scheme.

\subsection{Construction of a High-dimensional Feature Space}
Consider a planar shape discretized using a grid of dimension $n_1 \times n_2$.
We construct a $30$-dimensional feature vector $f^{x,y} \in \mathbb{R}^{30}$ for each shape pixel $(x,y)$ where $1 \le x \le n_1$ and $1 \le y \le n_2$.
In order to compute the feature component at each slot $k$ for $k = 1,2,\dots,30$, we first solve a linear system of equations in which the feature value of each shape pixel $u^{x,y}_k$ is related with the feature values of its four-neighboring pixels via (\ref{eqn:spe}) and we then normalize the obtained values as in (\ref{eqn:spenorm}).
\begin{align}
&\begin{aligned}
\label{eqn:spe}
{u^{x,y}_k} = \frac{{\rho_k}^2}{4 {\rho_k}^2 + 1} \, \big( 1 + {u^{x-1,y}_k} + {u^{x+1,y}_k} + {u^{x,y-1}_k} + {u^{x,y+1}_k} \big)
\end{aligned}\\
&\begin{aligned}
\label{eqn:spenorm}
{f^{x,y}_k} = \frac{u^{x,y}_k}{\max\limits_{1 \leq x' \leq n_1, 1 \leq y' \leq n_2}{u^{x',y'}_k}}
\end{aligned}
\end{align}
$u^{x,y}_k$ is solved only for the shape pixels and, it is considered $0$ for the pixels outside the shape.
$\rho_k$ is a scalar parameter which controls the dependence between the feature value of each shape pixel and the feature values of its four-neighbors.

\begin{figure}[H]
	\centering
	\includegraphics[width=0.3\linewidth]{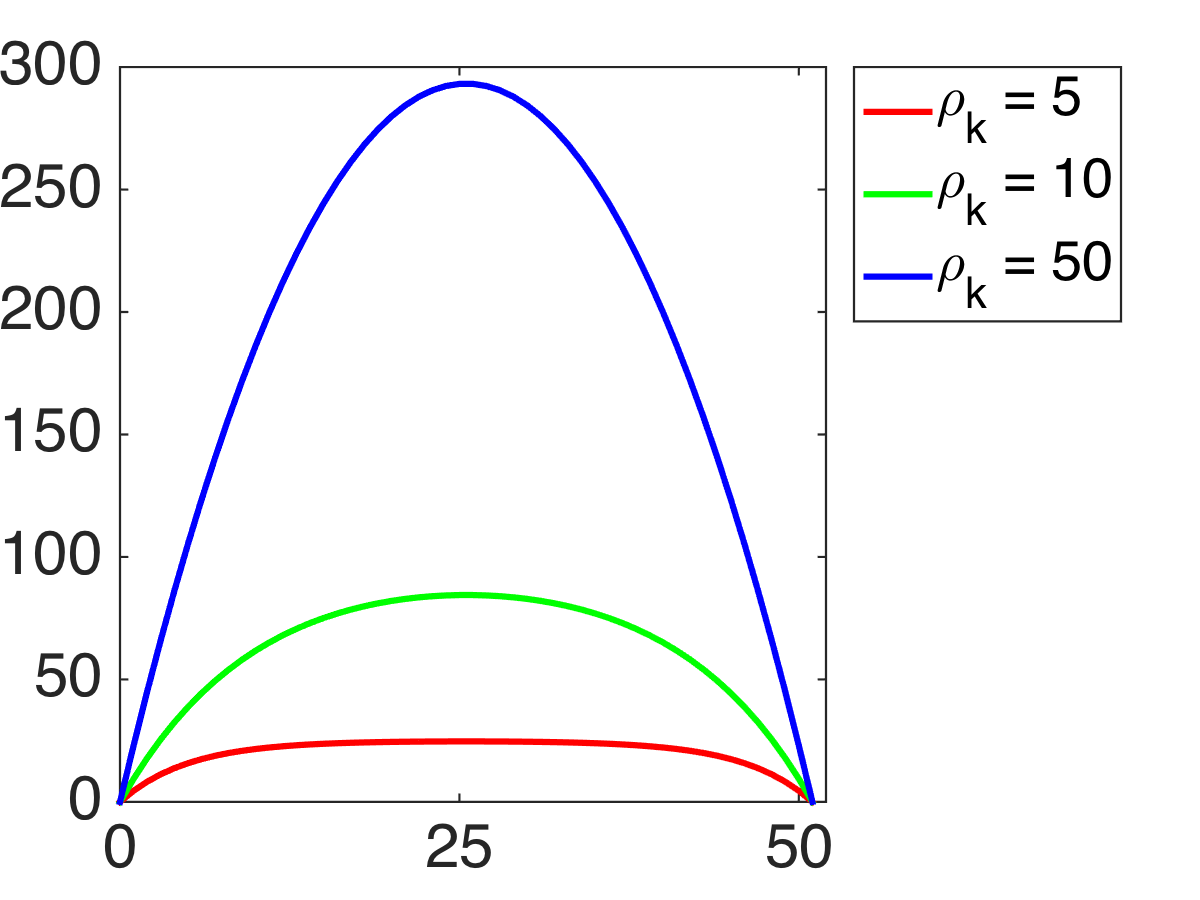}
	\includegraphics[width=0.3\linewidth]{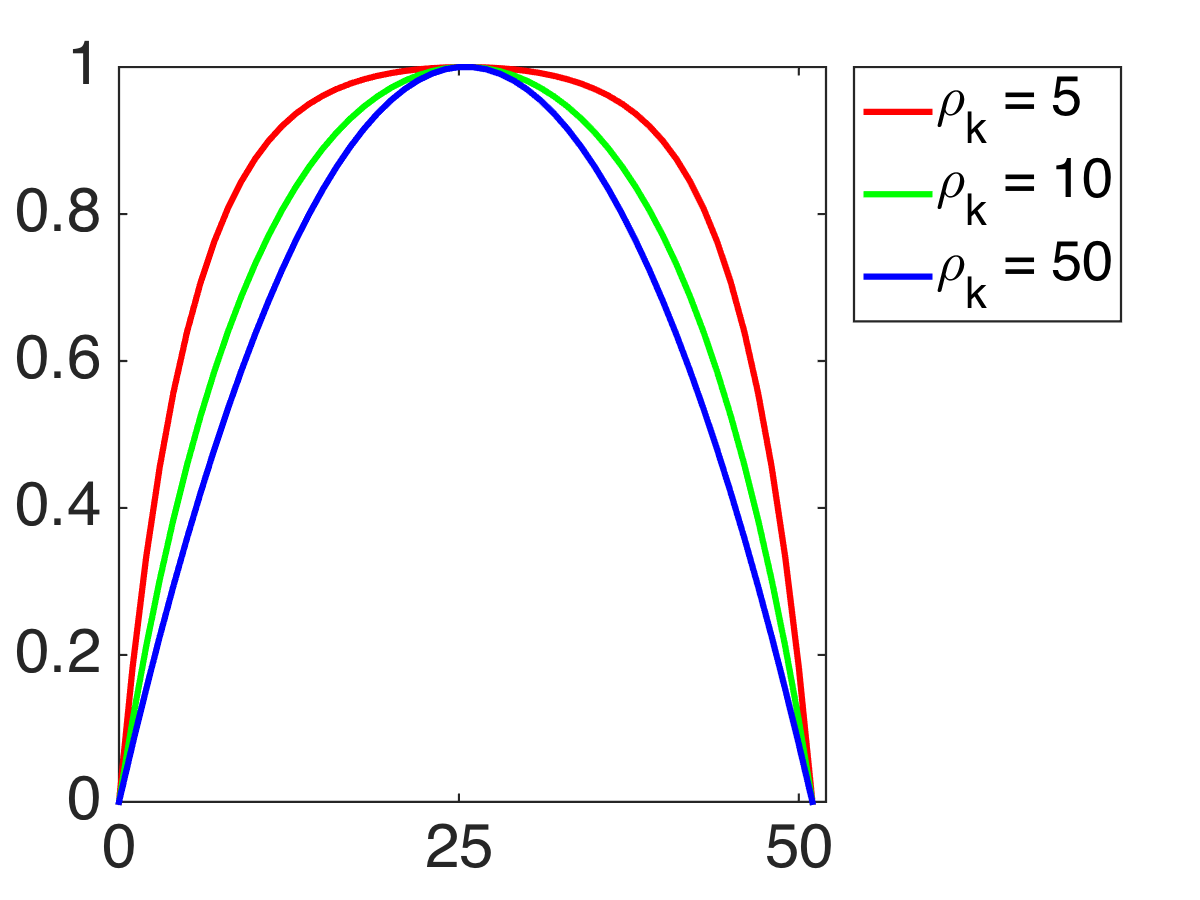}
	\caption{\label{fig:1} Features computed for a one-dimensional signal using three different values of the parameter $\rho_k$ (\textbf{left}) and normalized features (\textbf{right}).}
\end{figure}

In Fig.~\ref{fig:1}~(left), we show the features computed for a one-dimensional signal using three different values of $\rho_k$ corresponding to $1/10$, $1/5$ and $1$ times the whole signal length.
We normalize each feature to have the maximum value of $1$ (see Fig.~\ref{fig:1}~(right)).
We observe that the feature values monotonically increase towards the center.

By varying $\rho_k$, we obtain a collection of features each encoding a different degree of local interaction between the shape pixels and their surroundings.
We determine $\rho_k$ for $k = 1,2,\dots,30$ as $\rho_k = k \times \rho^{\star}/30$ where $\rho^{\star}$ represents the extent of the maximum interaction between the shape pixels and their surroundings.
In order to represent different shapes in a common feature space, we determine $\rho^{\star}$ for each shape individually as a measurement of the same shape property.

\subsection{Determining Multiple High-dimensional Feature Spaces}
We utilize two different shape measurements which are related with thickness of the shape body and the maximum distance between the shape extremities.
The first measurement $R$ is computed as the maximum value of the shape's distance transform which gives the distance of each shape point from the nearest boundary.
The second measurement $G$ is computed as the maximum value of the pairwise geodesic distances between the boundary points where the geodesic distance between a pair of points depends on the shortest path connecting them through the shape domain.

\begin{figure}[t]
	\centering
	\includegraphics[width=0.75\linewidth]{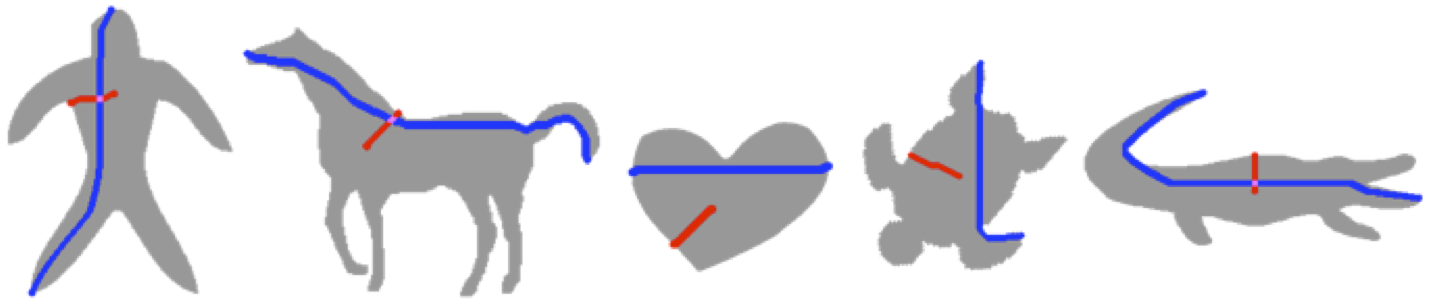}
	\caption{\label{fig:2} 
		$R$ and $G$ correspond to the thickness of the shape body (red) and the maximum distance between the shape extremities (blue), respectively.}
\end{figure}

As shown in Fig.~\ref{fig:2}, $R$ and $G$ provide characteristic shape information which can be used to define the extent of the local interactions between the shape pixels during the feature space construction. 
We construct six different feature spaces for which $\rho^{\star}$ is selected as multiples of $R$ or $G$, namely, $2 R$, $3 R$, $4 R$, $(2/3) G$, $(2/4) G$ and $(2/5) G$.

\subsection{Computing Distinctness of Shape Pixels via Each Feature Space}
We organize the feature vectors in the form of a matrix $D \in \mathbb{R}^{m{\times}30}$ where each row represents the feature vector computed for a shape pixel and $m$ is the total number of shape pixels.
The matrix $D$ is decomposed into a low-rank matrix $L$ and a sparse matrix $S$ via RPCA, which seeks to solve the following convex optimization problem:
\begin{equation}
\label{eq:rpca}
\min_{L, \, S \, \in \, \mathbb{R}^{m\times 30}} {||L||}_* + \lambda \, {||S||}_1 \;\;\; \textnormal{such that} \;\;\; L+S=D
\end{equation}
where ${||.||}_*$ denotes the sum of the singular values of the matrix, ${||.||}_1$ is the sum of the absolute values of all matrix entries and $\lambda$ is the weight of penalizing denseness of the sparse matrix $S$.
Various algorithms are proposed to solve the optimization problem in~(\ref{eq:rpca}). 
We use the inexact augmented Lagrange multipliers method for RPCA \cite{LinChenMa2010}, which is efficient and highly accurate.
We choose $\lambda = 1/\sqrt{m}$ as suggested by the available implementation of \cite{LinChenMa2010}.

The correlation between the feature vectors hence the shape pixels is encoded by the matrix $L$ whereas their discrimination is contained in the matrix $S$.
Thus, we define the distinctness of each shape pixel as the norm of the corresponding vector in the matrix $S$.
The shape pixels whose feature components vary more are found to be more distinct.
The shape articulations are associated with larger distinctness since they are thinner compared to the shape body and the constant value coming from the shape boundary is propagated faster in these regions during the feature computation.

\subsection{Partitioning Shapes into a Set of Regions via Each Feature Space}
We utilize the afore-mentioned property of the distinctness values in order to partition shapes into a set of regions.
We first divide the shape domain into two disjoint sets by thresholding at the mean distinctness value.
We further partition each set into multiple regions by dilating the two sets one after another in descending distinctness order.
In this way, we remove the connections between different regions of each set.
Radius of the structuring element used for dilating each pixel is determined using the distance of the pixel from the nearest boundary.

\subsection{Measuring Pairwise Shape Dissimilarity via Each Feature Space}
We describe each shape region by the normalized probability distribution of the distinctness values of its constituent pixels where the normalization is performed by making the probability sum equal to the ratio of the region area to the total shape area.
In order to estimate the probability distribution, we simply utilize the histogram of the distinctness values with a constant bin size $0.01$.
The dissimilarity between a pair of shapes is defined as the cost of the optimal assignment between their regions.
We use Hungarian matching for solving the optimal assignment problem.
We do not assume any relation between the regions of each shape.
Hungarian matching aims to find a one-to-one correspondence between the regions of the two shapes leaving some regions unmatched.
The cost of assigning two regions is simply taken as the sum of the absolute value of the difference between their normalized probability distributions.
The cost of leaving a region unmatched is taken as the sum of its normalized probability distribution, which is equal to the ratio of its area.

\subsection{Combining Pairwise Shape Dissimilarities Deduced from Multiple Feature Spaces}
In order to define the final dissimilarity of a pair of shapes, we compute a weighted average of the dissimilarities deduced from the six feature spaces.
The weight is $1/4$ for each of the dissimilarities via the feature spaces constructed using $R$ whereas it is $1/12$ for each of the dissimilarities via the feature spaces constructed using $G$.
The non-uniform weighting is due to that $R$ is more reliable than $G$ since the shape body is a more stable structure compared to the articulations.

\section{Experimental results}
\begin{figure}[H]
	\centering{\includegraphics[width=0.9\linewidth]{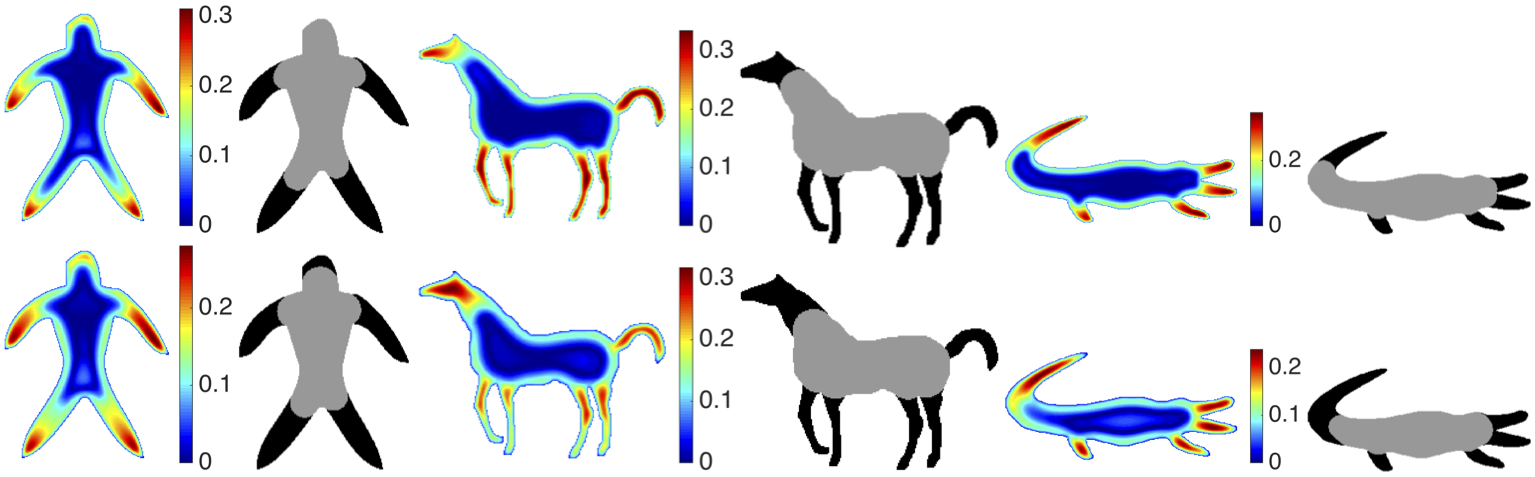}}
	\vspace*{-3mm}
	\caption{\label{fig:3}Distinctness values (color coded) and the corresponding partitioning results (gray vs. black) for three shapes via the feature spaces constructed for $\rho^\star = 3R$ (\textbf{top row}) and $\rho^\star = (2/4)G$ (\textbf{bottom row}).}
\end{figure}

As shown in Fig.~\ref{fig:3}, the distribution of distinctness values vary considering representations of different shapes via the same feature space or representations of a single shape via different feature spaces.
Grouping of the distinctness values on the shape domain provides partitioning of the shape into meaningful regions such as the shape body (gray) and the articulations (black) via simple operations.

\begin{figure}[t]
	\centering{\includegraphics[width=0.8\linewidth]{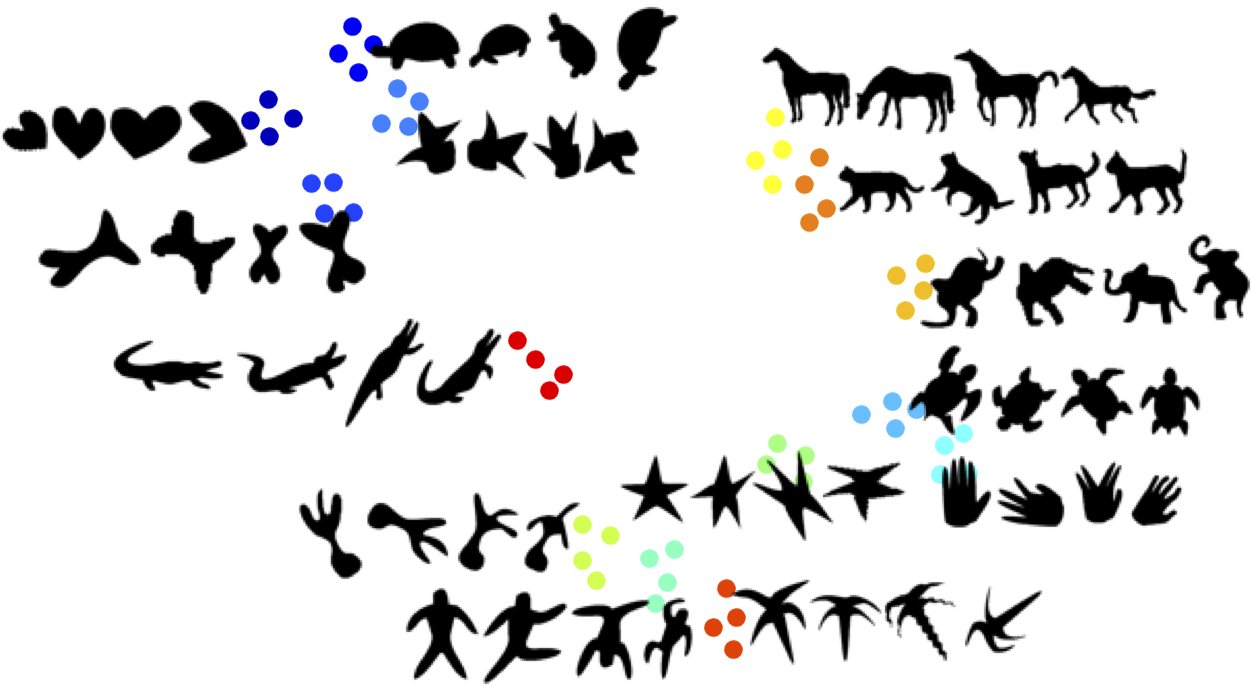}}
	\vspace*{-3mm}
	\caption{\label{fig:4}t-SNE mapping of the shapes from 56shapes dataset using the proposed dissimilarity measure.}
\end{figure}

In order to observe the clustering effect implied by the proposed dissimilarity measure, we utilize t-Distributed Stochastic Neighbor Embedding (t-SNE) \cite{tsne2008} which aims to map objects into a plane based on their pairwise dissimilarities.
In Fig.~\ref{fig:4}, we show the t-SNE mapping result for 56shapes \cite{Aslan2005} dataset which consist of $14$ shape categories each with $4$ shapes where the within category variations are due to transformations such as rotation, scaling and deformations of articulations.
We see that the shapes from the same category cluster together and the shapes from the similar categories (e.g. horse and cat shapes) are close to each other.

We compare our clustering results with state of the art methods using Normalized Mutual Information (NMI).
NMI measures the degree of agreement between the ground-truth category partition and the obtained clustering partition by utilizing the entropy measure.
The formulation of NMI is as follows \cite{csd2013}.
Let $n_i^j$ denote the number of shapes in cluster $i$ and category $j$, $n_i$ denote the number of shapes in cluster $i$, and $n^j$ denote the number of shapes in category $j$. Then NMI can be computed as follows:
\begin{equation}
\frac{2 \sum_{i = 1}^{I} \sum_{j = 1}^{J} \left(n_i^j/N\right) \log \left( \frac{\left(n_i^j/N\right)}{\left(n_i/N\right) \, \left(n^j/N\right)}\right)}{-\sum_{i = 1}^{I}\left(n_i/N\right) \log \left(n_i/N\right) -\sum_{j = 1}^{J}\left(n^j/N\right) \log \left(n^j/N\right)}
\end{equation}
where $I$ is the number of clusters, $J$ is the number of categories and $N$ is the total number of shapes.

A high value of NMI indicates that the obtained clustering matches well with the ground-truth category partition.
In order to compute NMI of our clustering result, we need to assign a cluster id to each shape.
Given the t-SNE mapping of a dataset obtained using our proposed dissimilarity measure, we apply affinity propagation \cite{ap2007} to partition the dataset into a number of clusters (which is chosen equal to the number of categories in the dataset).

In Table~\ref{tab:1}, we present NMIs of our proposed method and other state of the art methods on 56shapes \cite{Aslan2005}, 180shapes \cite{Aslan2008} and 1000shapes~\cite{baseski2009} datasets.
180shapes dataset consist of $30$ categories each with $6$ shapes.
1000shapes dataset consist of $50$ categories each with $20$ shapes.
%
%56shapes and 180shapes are highly articulated datasets compared to 1000shapes which contains unarticulated shape categories such as the face category.
%
The method of common structure discovery (CSD) \cite{csd2013} employs hierarchical clustering in which a common shape structure is constructed each time two clusters are merged into a single cluster where building a common shape structure requires matching skeleton graphs.
The method (skeleton path+spectral) presented in the work \cite{Bai2016} combines the skeleton path distance \cite{Bai2008} with spectral clustering.
The performance of these two skeleton-based methods decreases for 1000shapes dataset which contains unarticulated shape categories such as face category.
For 1000shapes dataset, the highest performance is obtained via the method (shape context+spectral) in \cite{Bai2016} which uses shape context descriptor \cite{sc2002}.
As the shape context descriptor is not robust to deformation of shape articulations, the performance decreases for highly articulated 56shapes and 180shapes datasets.
Inner distance shape context (IDSC) descriptor \cite{idsc2007} is an articulation invariant alternative to the shape context descriptor.
In the work \cite{csd2013}, the performance of IDSC combined with normalized cuts (Ncuts) algorithm is reported for the three datasets.
Overall, we accurately cluster the shapes from 56shapes dataset and our proposed method has the highest NMI average over the three datasets.
We observe that without constructing and matching graphs of shape components, our method performs comparable to the structural methods.

\begin{table}[H]
	\centering
	\caption{\label{tab:1}The clustering performance comparison using NMI.}
	\begin{tabular}{lllll}\toprule
			& 56shapes & 180shapes & 1000shapes & average\\\midrule
			CSD & 0.9734	 & 0.9694 & 0.8096 & 0.9175\\
			IDSC+Ncuts & 0.5660	 & 0.5423 & 0.5433 & 0.5505\\
			Shape context+spectral & 0.9418 & 0.9264 & \textbf{0.9676} & 0.9453\\
			Skeleton path+spectral & 0.9426 & \textbf{0.9746} & 0.9154 & 0.9442\\
			Proposed method & \textbf{1.0000} & 0.9651 & 0.9172 & \textbf{0.9608}
			\\\bottomrule
	\end{tabular}{}
\end{table}

\section{Summary and Conclusion}

We presented a novel representation scheme which does not involve any relational/structural model of the shape components.
Our representation scheme is based on a pixel-wise distinctness measure which is obtained by applying RPCA to the shape pixels represented in a high-dimensional feature space.
The distinctness measure is proven to be very useful.
Its spatial distribution on the shape domain provides easy partitioning of the shape into meaningful regions and its probability distribution provides a description of each region.
We define a pairwise dissimilarity measure as the cost of optimal mapping between regions of the shapes.
The results of the clustering experiments on highly articulated shape datasets show that our proposed method performs comparable to state of the art methods.

%%%%%%%%%%%%%%%%%%%%%%%%%%%%%%%%%%%%%%%%%%
\vspace{6pt} 

%%%%%%%%%%%%%%%%%%%%%%%%%%%%%%%%%%%%%%%%%%
%% optional
%\supplementary{The following are available online at \linksupplementary{s1}, Figure S1: title, Table S1: title, Video S1: title.}

% \supplementary{The following are available at \linksupplementary, Figure S1: title, Table S1: title, Video S1: title. A supporting video article is available at doi: link.}

%%%%%%%%%%%%%%%%%%%%%%%%%%%%%%%%%%%%%%%%%%
\authorcontributions{A.G. and S.T. contributed to the design and development of the proposed method and to the writing of the manuscript. A.G. contributed additionally to the software implementation and testing of the proposed method.}

%%%%%%%%%%%%%%%%%%%%%%%%%%%%%%%%%%%%%%%%%%
\funding{This research was funded by TUBITAK grant number 112E208.}

%%%%%%%%%%%%%%%%%%%%%%%%%%%%%%%%%%%%%%%%%%
%\acknowledgments{In this section you can acknowledge any support given which is not covered by the author contribution or funding sections. This may include administrative and technical support, or donations in kind (e.g. materials used for experiments).}

%%%%%%%%%%%%%%%%%%%%%%%%%%%%%%%%%%%%%%%%%%
\conflictsofinterest{The authors declare no conflict of interest. The funding sponsors had no role in the design of the study; in the collection, analyses, or interpretation of data; in the writing of the manuscript, and in the decision to publish the results.} 

%%%%%%%%%%%%%%%%%%%%%%%%%%%%%%%%%%%%%%%%%%
%% optional
\abbreviations{The following abbreviations are used in this manuscript:\\

\noindent 
\begin{tabular}{@{}ll}
2D & Two-dimensional\\
RPCA & Robust Principal Component Analysis\\
t-SNE & t-Distributed Stochastic Neighbor Embedding\\
NMI & Normalized Mutual Information\\
CSD & Common Structure Discovery\\
IDSC & Inner Distance Shape Context\\
Ncuts & Normalized Cuts
\end{tabular}}

%%%%%%%%%%%%%%%%%%%%%%%%%%%%%%%%%%%%%%%%%%
%% optional
%\appendixtitles{no} %Leave argument "no" if all appendix headings stay EMPTY (then no dot is printed after "Appendix A"). If the appendix sections contain a heading then change the argument to "yes".
%\appendixsections{multiple} %Leave argument "multiple" if there are multiple sections. Then a counter is printed ("Appendix A"). If there is only one appendix section then change the argument to "one" and no counter is printed ("Appendix").
%\appendix
%\section{}
%\subsection{}
%The appendix is an optional section that can contain details and data supplemental to the main text. For example, explanations of experimental details that would disrupt the flow of the main text, but nonetheless remain crucial to understanding and reproducing the research shown; figures of replicates for experiments of which representative data is shown in the main text can be added here if brief, or as Supplementary data. Mathematical proofs of results not central to the paper can be added as an appendix.

%\section{}
%All appendix sections must be cited in the main text. In the appendixes, Figures, Tables, etc. should be labeled starting with `A', e.g., Figure A1, Figure A2, etc. 


%=====================================
% References, variant A: internal bibliography
%=====================================
\reftitle{References}
\end{document}